\documentclass{article}
\usepackage{microtype}
\usepackage{enumitem}
\usepackage{amsmath}
\usepackage{hyperref}
\usepackage{framed}
\usepackage{subfigure}
\usepackage{booktabs} 
\usepackage{float}
\usepackage{multirow} 
\usepackage{array} 
\usepackage{graphicx} 
\usepackage[table]{xcolor} 
\usepackage{tikz}
\usepackage{amssymb}
\usepackage{pifont}
\usepackage{fontawesome}     

\usepackage{hyperref}

\definecolor{darkgreen}{RGB}{0,100,0} %

\newcommand{\cmark}{\textcolor[rgb]{0,0.5,0}{\checkmark}} 
 
\usepackage[accepted]{mlsys2025}

\mlsystitlerunning{}
\hypersetup{
    colorlinks=true, 
    linkcolor=red,   
    citecolor=blue   
}
\begin{document}

\twocolumn[
\mlsystitle{\textsc{Infant Agent}: A Tool-Integrated, Logic-Driven Agent with Cost-Effective API Usage}



\mlsyssetsymbol{equal}{*}

\begin{mlsysauthorlist}
\mlsysauthor{Bin Lei}{one}
\mlsysauthor{Yuchen Li}{two}
\mlsysauthor{Yiming Zeng}{one}
\mlsysauthor{Tao Ren}{three}
\mlsysauthor{Yi Luo}{one}
\mlsysauthor{Tianyu Shi}{four}
\mlsysauthor{Zitian Gao}{five}
\mlsysauthor{Zeyu Hu}{one}
\mlsysauthor{Weitai Kang}{six}
\mlsysauthor{Qiuwu Chen}{two}
\\
$^1$University of Connecticut \quad $^2$AIGCode \quad $^3$TikTok \\
$^4$University of Toronto \quad $^5$University of Sydney \quad $^6$Illinois Institute of Technology \\
\end{mlsysauthorlist}



\mlsyskeywords{Machine Learning, MLSys}

\vskip 0.3in

\begin{abstract}
Despite the impressive capabilities of large language models (LLMs), they currently exhibit two primary limitations, \textbf{\uppercase\expandafter{\romannumeral 1}}: They struggle to \textbf{autonomously solve the real world engineering problem}. \textbf{\uppercase\expandafter{\romannumeral 2}}: They remain \textbf{challenged in reasoning through complex logic problems}. 
To address these challenges, we developed the \textsc{Infant Agent}, integrating task-aware functions, operators, a hierarchical management system, and a memory retrieval mechanism. Together, these components enable large language models to sustain extended reasoning processes and handle complex, multi-step tasks efficiently, all while significantly reducing API costs. Using the \textsc{Infant Agent}, GPT-4o's accuracy on the SWE-bench-lite dataset rises from $\mathbf{0.33\%}$
 to $\mathbf{30\%}$, and in the AIME-2024 mathematics competition, it increases GPT-4o's accuracy from $\mathbf{13.3\%}$ to $\mathbf{37\%}$.
\end{abstract}
]




\section{Introduction}

LLMs have achieved remarkable advancements across various domains, primarily due to their powerful pattern recognition and contextual understanding capabilities~\cite{naveed2023comprehensive}. Trained on extensive datasets, LLMs can generate coherent and high-quality outputs, tackle complex tasks, and demonstrate strong adaptability across a wide range of applications~\cite{yang2024harnessing}. However, despite their impressive capabilities, LLMs still face two significant limitations~\cite{hadi2024large}: 
\begin{enumerate}[label=\textbf{\arabic*.}]
    \item \textit{\textbf{LLMs themselves struggle with interaction with the physical world, which limits their capability to autonomously address certain engineering problems.}}
    \item \textit{\textbf{LLMs often struggle with multi-step logical reasoning, which limits their ability to solve complex logic problems and hinders their capacity for innovation.}}
\end{enumerate}
To mitigate these limitations and further unlock the potential of LLMs, we developed the \textsc{Infant Agent}. It is a fully autonomous, multi-agent workflow that integrates step-by-step reasoning, tool invocation, environment interaction, feedback adjustments, and evaluation summaries. Each step in this process is autonomously determined by the Agent itself. The entire workflow begins with the user's input and then enters an infinite loop, where the Agent autonomously determines every step. The process continues until the Agent concludes that the task is complete or the system reaches its budget restrictions .

\begin{figure}[ht]
    \centering
    \includegraphics[width=0.48\textwidth]{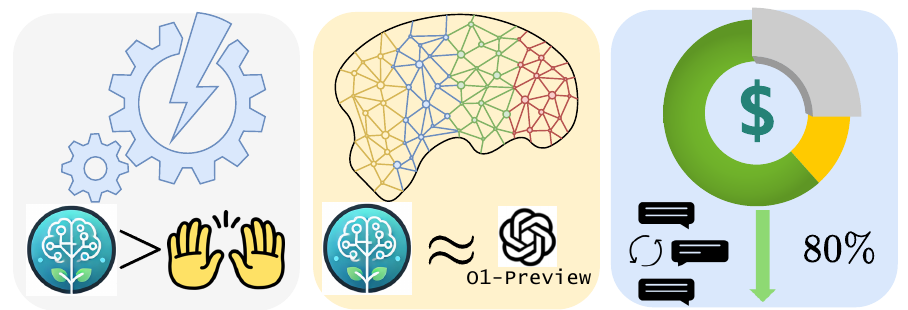} 
    \caption{
Overall performance summary. \includegraphics[height=0.7em]{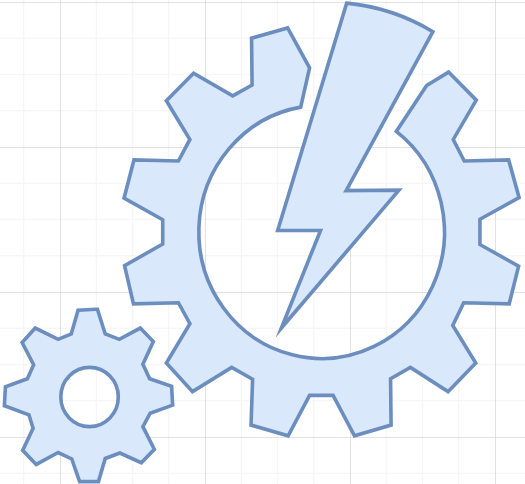}: Real world engineering problem. \includegraphics[height=0.7em]{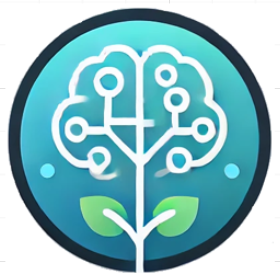}: \textsc{Infant Agent}. \includegraphics[height=0.7em]{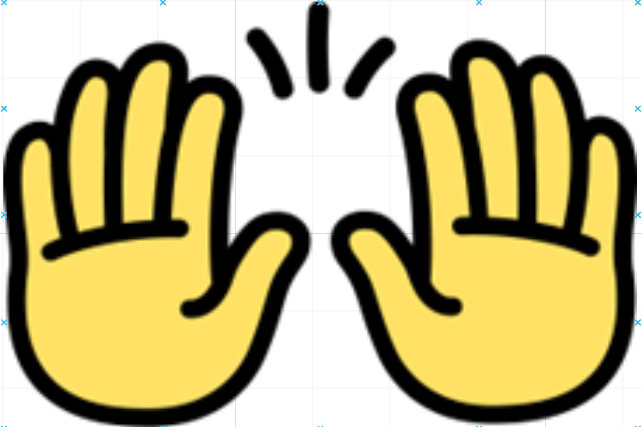}: OpenHands CodeActAgent. \includegraphics[height=0.7em]{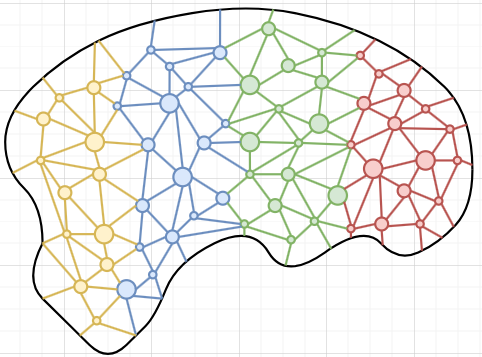}: Logic Reasoning problem. \includegraphics[height=0.7em]{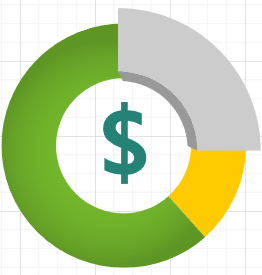}: API token cost. \includegraphics[height=0.7em]{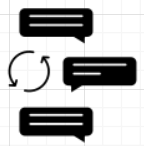}: API token.} 
    \label{fig:introduction} 
\end{figure}
In Figure~\ref{fig:introduction}, we summarize the overall performance of the \textsc{Infant Agent}. By invoking tools and performing file edits to solve real-world engineering problems using GPT-4o~\cite{OpenAI_GPT4o}, \textsc{Infant Agent} surpasses OpenHands~\cite{wang2024opendevin} CodeActAgent~\cite{wang2024executable} v$1.8$ on SWE-Bench-Lite~\cite{jimenez2023swe} by $8$ percentage points ($30\%$ vs $22\%$).
For complex logical reasoning problems, with GPT-4o + Qwen2.5-72B-Instruct~\cite{Qwen2.5_72B_Instruct}, \textsc{Infant Agent} achieves the same Pass@1 accuracy as o1-preview~\cite{OpenAI_Learning_to_Reason} on the AIME2024~\cite{AIME2024} dataset ($37\%$ vs $37\%$). Additionally, \textsc{Infant Agent} segments memory modules for different tasks and extracts memory in various situations, resulting in a nearly $80\%$ reduction in API token cost.

In summary, our main contributions are as follows:
\begin{enumerate}[label=\textbf{\arabic*.}]
    \item \textit{\textbf{We developed \textsc{Infant Agent}}, which not only can invoke tools to solve real-world engineering problems but also engages in logical reasoning and self-reflection.}
    \item \textit{\textbf{We proposed a hierarchical agent collaboration system}, which mitigates issues of ineffective outputs caused by an excessive number of built-in commands or overly long few-shot examples.}
    \item \textit{\textbf{We implemented a memory retrieval mechanism}, which reduces API token costs by nearly 80\% compared to using the full memory for inference each time.
}
\end{enumerate}
\section{Related Work}
The use of agents has become increasingly important as a means to automate and optimize complex tasks, particularly those that involve multi-step processes or require interaction with external resources~\cite{qian2024chatdev,abbasian2023conversational}. Agents offer the potential to enhance efficiency, reduce human intervention, and manage intricate workflows~\cite{buhler2005towards, yan2001integration}. 

\textbf{Agents for General Task:} \textit{AutoGPT}~\cite{AutoGPT}, \textit{BabyAGI}~\cite{BabyAGI}, \textit{AgentGPT}~\cite{AgentGPT}, and \textit{AutoGen}~\cite{Microsoft_AutoGen} are designed to tackle a broad range of general tasks by breaking down user queries into actionable components. These agents typically perform operations such as decomposing user questions, browsing online resources, and providing feedback. Among these, \textit{AutoGPT} distinguishes itself by autonomously linking to external sources, while \textit{AgentGPT} requires user input for certain steps, ensuring user-guided interactions. \textit{AutoGen}, on the other hand, supports collaboration among multiple agents, facilitating more efficient task execution through cooperative problem-solving.

\textbf{Software Automation Agents:} \textit{MetaGPT}~\cite{hong2023metagpt} and \textit{Aider}~\cite{Aider} focus on automating the software development pipelines. Both follow a structured cycle of writing, running, debugging, and refining code. \textit{MetaGPT} is designed specifically for end-to-end automated software development, offering an integrated solution for code generation and testing. In contrast, \textit{Aider} assists developers by auto-completing code, identifying bugs, and providing optimizations within the user's daily workflow, making it suitable for enhancing productivity in practical development scenarios. \textit{Devin}~\cite{Devin}, \textit{OpenHands}, and \textit{SWE-Agent}~\cite{yang2024swe} are specialized in managing complex code file operations, demonstrating the capability to solve real-world code issues, such as those encountered on GitHub. These agents are tailored to handle intricate file manipulation tasks and interact with large codebases, showcasing their utility in software maintenance.

By leveraging specialized agents, these approaches seek not only to automate routine and complex workflows but also to enhance the efficiency and scalability of task execution. They empower users to handle sophisticated problem-solving requirements across various domains, from software development to scientific research. The Agents’ abilities to manage, prioritize, and execute multiple tasks simultaneously allows for streamlined operations, leading to optimized performance and enabling users to achieve higher productivity and accuracy in diverse environments.
\section{Infant Agent}

\subsection{Overall Architecture}
\begin{figure*}[t]
    \centering
    \includegraphics[width=\textwidth]{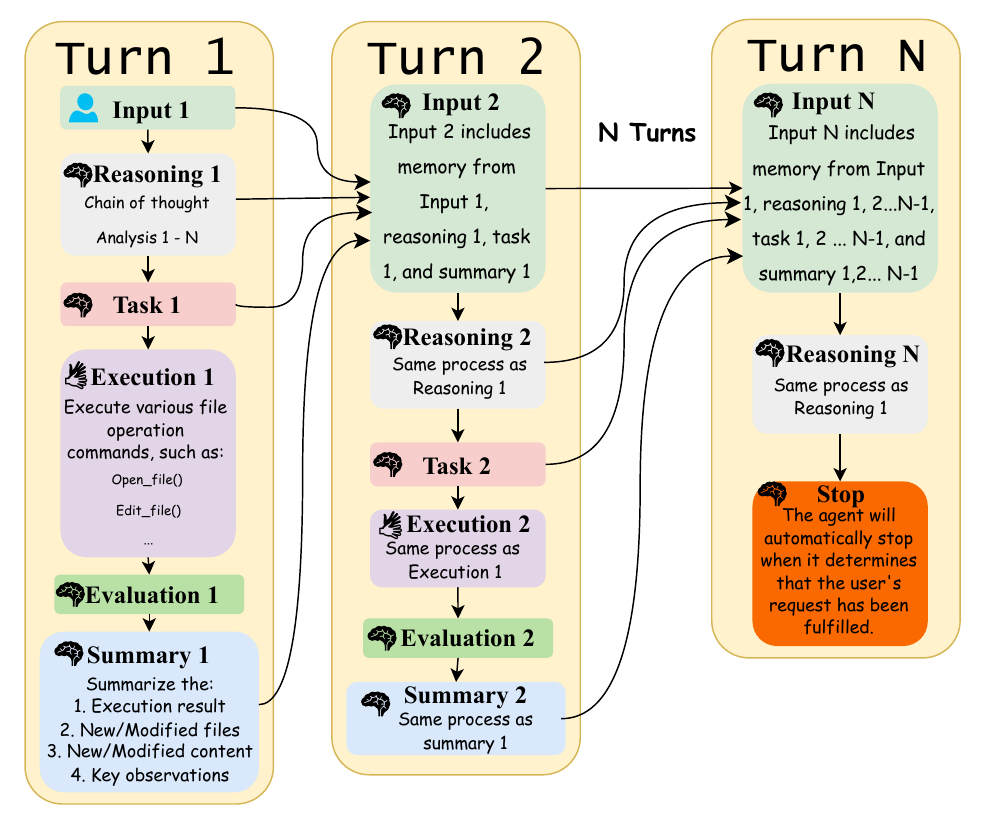} 
    \caption{The overall architecture of \textsc{Infant Agent.} \includegraphics[height=0.7em]{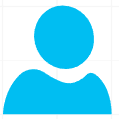}: Request from the User. \includegraphics[height=0.7em]{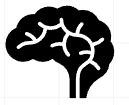}: Brain level agent. \includegraphics[height=0.7em]{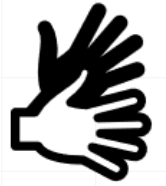}: Hand level agent.} 
    \label{fig:Overall_Architecture} 
\end{figure*}
In 
Figure~\ref{fig:Overall_Architecture}, we illustrate the overall architecture of \textsc{Infant Agent}. All its operations can be contained within an infinite loop. In each loop, as long as it determines that the user’s request has not yet been fulfilled, it will sequentially perform the following actions: reasoning and analysis, task scheduling, task execution, results evaluation, and summarization. Except for the task execution step, which is handled by the \textit{hand}-level Agent, all other operations are managed by the \textit{brain}-level Agent. In the next loop iteration, the actions executed by the \textit{brain}-level Agent are summarized in the form of dialogue history and used as input for the next iteration. The entire process is fully automated by the \textsc{Infant Agent}, and once it determines that the user’s request is complete, the infinite loop will break.
The specific functionalities of each operation are introduced as follows:

\textbf{\textit{Input:}}
We parse the user's input and extract a \textit{mandatory requirement} based on the user's request. In subsequent operations, we continually remind the Agent that its response must satisfy this \textit{mandatory requirement}. This acts as a constraint for the Agent, thereby encouraging it to respond aligned with the user's expectations. The \textit{mandatory requirements} vary depending on the scenario. For example, in coding tasks, the \textit{mandatory requirement} might be unit tests, while in writing tasks, it could be writing preferences.

\textbf{\textit{Reasoning:}}
By default, we employ the conventional chain-of-thought~\cite{wei2022chain} approach for reasoning. Each time, the Agent is required to only return one step of analysis. Additionally, similar to previous works, for complex problems, the reasoning process may trigger multi-round voting or reflection.

\textbf{\textit{Task:}}
If the Agent determines that a task needs to be executed, the \textit{brain}-level Agent will assigns it to the \textit{hand}-level Agent. The output of this step includes specific task objective, detailed steps, and expected outcome. For example, in a coding task, if the \textit{brain}-level Agent wants to debug in the file \texttt{test.py}, the task description would be: 
\begin{framed}
\textit{Please open the file \texttt{test.py}, add a \texttt{print} statement at line ..., and then run \texttt{python test.py} to provide me with the output. I expect to see the value of the variable ... printed.}
\end{framed}

\textbf{\textit{Execution:}}
The execution of specific tasks is handled by the \textit{hand}-level Agent. We follow the task execution methods used by OpenHands~\cite{wang2024opendevin} and SWE-Agent~\cite{yang2024swe}. The \textit{hand}-level Agent generates demand-matching file operation functions based on the task requirements, which are then invoked within a sandbox environment. These functions return the specific results obtained after execution.

\textbf{\textit{Evaluation:}}
After execution, the \textit{brain}-level Agent evaluates the returned results by comparing them to the expected output specified in the task requirements. This process allows the agent to assess the accuracy and completeness of the executed actions. If the brain-level agent determines the output meets the task requirements, the task is marked as completed. If not, the agent is asked to retry with a different solution considering the wrong results and try to correct the errors, thereby ensuring the task meets quality standards before proceeding to subsequent steps.

\textbf{\textit{Summary:}}
The summarization process is essentially a compression of the execution steps and results from the current turn. It mainly records the execution outcome, file modifications, and key conclusions. We standardize all file additions, deletions, reads, and modifications using \texttt{git patches}. This approach not only ensures accurate tracking of file changes but also reduces token usage and allows for modification of these patches at any time using git commands.

\textbf{\textit{Stop:}}
Once the Agent determines that the user's request has been resolved, it compares the final result with the \textit{mandatory requirement} formalized from the input. If the result meets the \textit{mandatory requirement}, the Agent will automatically execute the exit command. Otherwise, the infinite loop continues until the result satisfies the requirement or the budget is run out.

For clarity, we have provided an example of \textsc{Infant Agent} solving a problem in the Appendix~\ref{Appendix:Pipeline_Demonstration_Example}.

\subsection{Hierarchical Agent Collaboration System}
\begin{figure}[ht]
    \centering
    \includegraphics[width=0.48\textwidth]{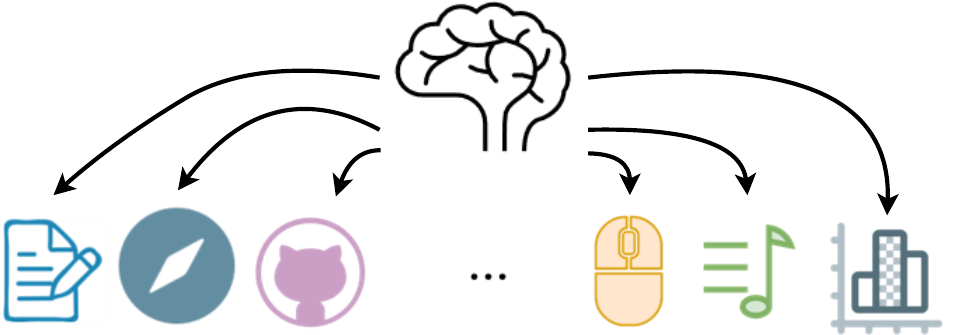} 
    \caption{Hierarchical Agent Collaboration System. \includegraphics[height=0.7em]{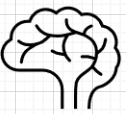}: Brain level agent. \includegraphics[height=0.7em]{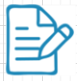}: File editor. \includegraphics[height=0.7em]{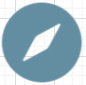}: Browser. \includegraphics[height=0.7em]{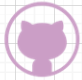}: Code agent. \includegraphics[height=0.7em]{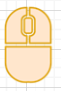}: Mouse/keyboard operation. \includegraphics[height=0.7em]{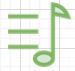}: Music. \includegraphics[height=0.7em]{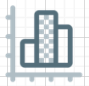}: Data analysis.} 
    \label{fig:Hierarchical_Agent} 
\end{figure}
One challenge in building an agent that can adapt to various situations is that as the number of command format prompts and few-shot examples increases in in-context learning, the instruction-selecting capability of LLMs tends to diminish. To address this issue, \textsc{Infant Agent} employs a hierarchical collaboration structure. As shown in Figure~\ref{fig:Hierarchical_Agent}, agents are divided into \textit{brain}-level and \textit{hand}-level agents. The \textit{brain}-level agents handle all the reasoning, while the \textit{hand}-level agents are responsible for executing tasks by invoking different tools, such as file editing, web browsing, and code execution. Each hand-level agent can be designed with prompts or trained on carefully curated datasets specifically tailored to its task, which not only significantly reduces token usage but also nearly eliminates all incorrect command invocations. For example, in our experiments, we tested a pure code task where browser-related commands were mixed with code commands. In $\mathbf{13.9\%}$ of cases, the agent incorrectly invoked browser-related commands. However, with the hierarchical structure, the percentage of incorrect browser command invocations dropped to $\mathbf{0\%}$. For detailed experimental information, see Experiment~\ref{exp:Error_Command_Correction}.

\subsection{Memory Retrieval Mechanism}
n Figure~\ref{fig:Memory}, we designed a memory retrieval mechanism to prevent excessive API cost consumption. The specific explanation for each step is as follows:
\begin{figure}[ht]
    \centering
\includegraphics[width=0.48\textwidth]{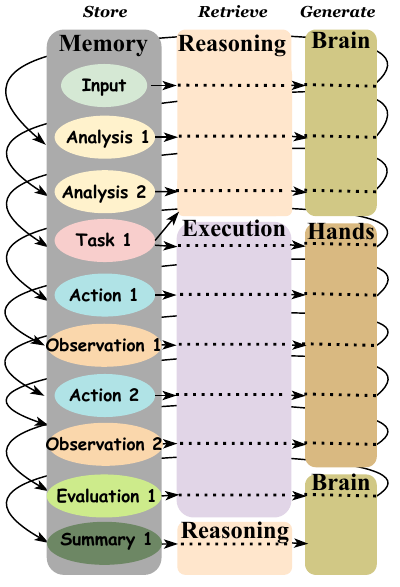} 
    \caption{Overview of Memory Storage, Retrieval, and Generation.} 
    \label{fig:Memory} 
\end{figure}

\textbf{\textit{Storage:}}
All responses (\texttt{string}) generated by the model will be parsed in a specific way and uniformly stored in historical memory as instances in sequence. For example, when we ask the model to analyze a coding problem, a sample response is as follows:
\begin{framed}
\textit{I will help you to analyze this problem.
To solve this problem, we need to merge multiple sorted linked lists into a single sorted linked list.}
\end{framed}

It will be parsed as a \texttt{Class}:
\begin{framed}
\texttt{class Analysis:}

~~~~\texttt{content: str = Merge multiple sorted linked lists into a single sorted linked list.} 
\end{framed}
This allows us to extract key information and apply it in various forms across different scenarios, such as evaluation and summarization, thereby prompting the model to generate responses in different formats accordingly.

\textbf{\textit{Retrieval:}}
Before generating a new response, the Agent's historical memory is retrieved. The reasoning and task execution parts are separated: \textit{Input, Analysis,} and {Summary} memories are retrieved during the Reasoning process; \textit{Action} and {Observation} memories are retrieved during the Execution process; and \textit{Task} memories are retrieved during both the Reasoning and Execution processes.
Memory retrieval is designed to help the Agent categorize memories, facilitating the assignment of tasks to different levels of Agents in the next step.

\textbf{\textit{Generation:}}
Closed-source models are used solely as the \textit{brain} of our Agent. For simple and repetitive tasks, smaller and cheaper, or open-source models are utilized for execution, serving as the \textit{hands} of the Agent.
\textit{Reasoning, summarization, and evaluation} tasks are managed by the brain-level agent, while task \textit{execution} is handled by the \textit{hand}-level agent. The \textit{execution} phase, particularly the observation step, consumes the most tokens since it involves reading files and other operations that contain a lot of textual information. By outsourcing the execution process to the \textit{hand}-level agent, we can significantly reduce reliance on expensive closed-source models.

\subsection{Token computational analysis}
 We perform a analysis of input tokens and output tokens before and after applying the memory extraction mechanism.
We assume that for each question, there are 
$n$ analyses per turn, 
$m$ action-observation pairs, and 
$k$ turns. Based on Figure~\ref{fig:Overall_Architecture}, we have:

The input tokens before applying the memory extraction:
\begin{multline*}
Token_{\text{in\_bef}} = k(3+2m+n) \, Token_{\text{input}} \\
+ Token_{\text{sumy}} \sum_{i=0}^{k-1} \big((k-i-1)(3+2m+n)\big) \\
+ Token_{\text{eval}} \sum_{i=0}^{k-1} \big((k-i)(3+2m+n) - n - 2m - 2\big) \\
+ Token_{\text{task}} \sum_{i=0}^{k-1} \big((k-i)(3+2m+n) - n - 1\big) \\
+ Token_{\text{analysis}} \sum_{i=0}^{k-1} \sum_{j=0}^{n-1} \big((3+2m+n)k - (3+2m)i - j - 1\big) \\
+ Token_{\text{action}} \sum_{i=0}^{k-1} \sum_{j=0}^{m} \big((3-i)(3+2m+n) - 2n - 2 - 2j\big) \\
+ Token_{\text{obs}} \sum_{i=0}^{k-1} \sum_{j=0}^{m} \big((3-i)(3+2m+n) - 2n - 2 - 2j - 1\big)
\end{multline*}

The output tokens before applying the memory extraction mechanism are:
\begin{multline*}
Token_{\text{out\_bef}} = k \times (Token_{\text{sumy}} + Token_{\text{eval}} + Token_{\text{task}}) \\
+ nk \times Token_{\text{analysis}}
+ mk \times (Token_{\text{action}} + Token_{\text{obs}})
\end{multline*}

The input tokens after applying the memory extraction mechanism are:
\begin{multline*}
Token_{in\_aft} =k(2+n) Token_{input} \\+Token_{sumy}\sum _{i=0}^{k-1}(( k-i)( 2+n) -n-2) \\+Token_{task}\sum _{i=0}^{k-1}(( k-i)( 2+n) -n-1)
\\+\ Token_{analysis}\sum _{i=0}^{k-1}\sum _{j=1}^{n}(( k-i)( 2+n) -j)
\end{multline*}

The output tokens after applying the memory extraction mechanism are:
\begin{multline*}
Token_{out\_aft}=k(Token_{sumy}+Token_{task}) \\
+ nk(Token_{analysis})
\end{multline*}

Where: $Token_{\text{input}}$ represents the token count of the request made by the user. $Token_{\text{sumy}}$ is the average token count for the \textit{summarization} step. $Token_{\text{eval}}$ refers to the average token count for the \textit{evaluation} step. $Token_{\text{task}}$ indicates the average token count for each \textit{task}, $Token_{\text{analysis}}$ denotes the average token count for each \textit{analysis} step, $Token_{\text{action}}$ represents the average token count for each \textit{action during execution}, and $Token_{\text{obs}}$ is the average token count for each \textit{observation step during execution}. 

According to the sampling of 100 different test cases, we obtained the average values of each variable as follows: $n = 2.53$,  $m=3.78$,  $k=5.64$, $Token_{\text{input}}=359$, $Token_{\text{sumy}}=784$, $Token_{\text{eval}}=7.54$, 
$Token_{\text{task}}=754$, $Token_{\text{analysis}}=148$, $Token_{\text{action}}=227$, $Token_{\text{obs}}=1994$. 
Substituting into the above formulas, we find that applying the memory extraction mechanism for a single user request can save $\mathbf{79.81\%}$ of input tokens and $\mathbf{83.06\%}$ of output tokens. This theoretical derivation aligns closely with the experimental test results, with specific experimental tests detailed in Experiment~\ref{exp:API_Token_Savings}.

\subsection{Execution Tools}
\begin{figure*}[t]
    \centering
\includegraphics[width=\textwidth]{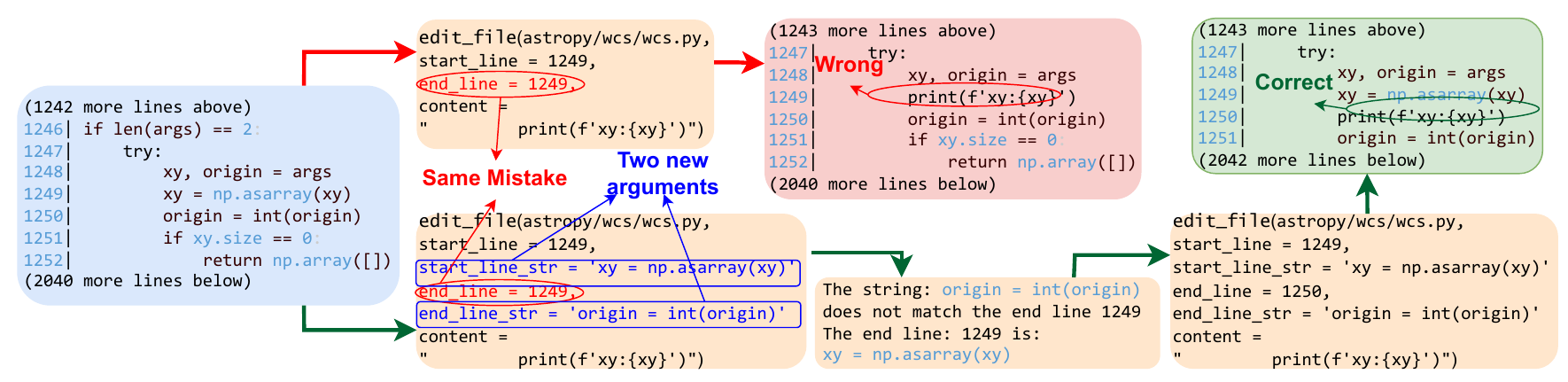} 
    \caption{Differences between the File-Editing commands of Infant-AI and SWE-Agent. \includegraphics[height=0.7em]{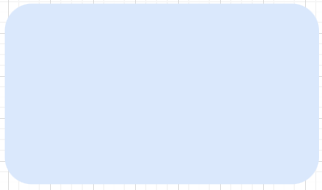}: Original file content. \includegraphics[height=0.7em]{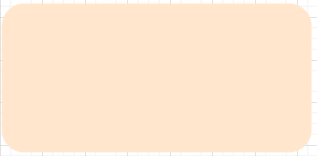}: Command generated by the Agent. \includegraphics[height=0.7em]{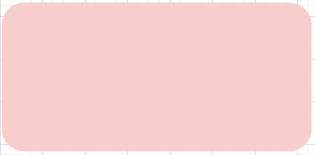}: Final modified file content generated by SWE-Agent. \includegraphics[height=0.7em]{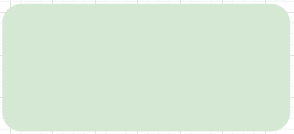}: Final modified file content generated by \textsc{Infant Agent}. \includegraphics[height=0.7em]{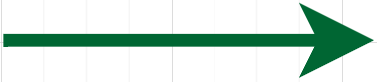}: Modification process of the file by \textsc{Infant Agent}. \includegraphics[height=0.7em]{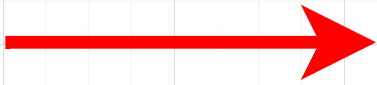}: Modification process of the file by SWE-Agent.} 
    \label{fig:edit_file} 
\end{figure*}
While LLMs are highly capable in natural language processing, our practical experiments show that even some of the most advanced models, like GPT-4o and Claude 3.5 Sonnet, frequently make fundamental mistakes, such as sequence misalignment, particularly when processing large files. This issue is not unique to individual models; many state-of-the-art agents, including Cursor~\cite{Cursor} and OpenHands, encounter similar challenges. These errors suggest that LLMs, despite their language strengths, still struggle with accuracy in detailed, sequence-dependent tasks within extensive datasets.

To address this issue, we first enhanced the original file-editing commands of OpenHands and SWE-Agent, with specific differences shown in Figure~\ref{fig:edit_file}. 
We added two new parameters, \texttt{start\_line\_string} and \texttt{end\_line\_string}, to the original SWE-Agent editing command \texttt{edit\_file(file\_name, start\_line, end\_line, new\_content)}. These parameters require that the line number of \texttt{start\_line} must correspond to \texttt{start\_line\_string}, and the line number of \texttt{end\_line} must match \texttt{end\_line\_string}. If they do not match, the Agent will automatically issue prompt commands to guide the LLM in adjusting the original \texttt{edit\_file()} command until it matches correctly. 

We made this improvement because we found that LLMs have a strong understanding of text but slightly less proficiency in discriminate numbers. As a result, they can often identify the correct editing location but may struggle with specifying the correct line number. This enhancement to the file-editing command improved the accuracy of SWE-Agent's file-editing function from $\mathbf{72.9\%}$ to $\mathbf{96.8\%}$. The specific experimental details are provided in Experiment~\ref{exp:File_Editing}.

Additionally, we customized two commands specifically for code tasks: \texttt{replace\_function(file\_name, function\_to\_replace, new\_code)}, which replaces a specific function in a given file based on its signature, and \texttt{Trace\_code\_switch(True/False)}, which enables the Agent to track essential functions called during execution, regardless of whether they run successfully or encounter errors. This tracing capability helps the Agent pinpoint potential issues in code logic by identifying functions where problems may arise.

\section{Experiment}
In this section, we tested \textsc{Infant Agent}'s performance across four key datasets: SWE-bench-lite~\cite{yang2024swe}, AIME2024, GPQA-diamond~\cite{rein2023gpqa}, and Codeforce contests~\cite{Codeforces}. SWE-bench-lite evaluates the agent's ability to address real-world engineering problems, while AIME2024 and Codeforce test its skills in handling complex logical tasks. Additionally, GPQA Diamond requires strong logical reasoning, autonomous online information retrieval, and code execution for calculations. In the ablation studies, we examined improvements from the Hierarchical Agent Collaboration System in command accuracy, token savings from the Memory Retrieval Mechanism, and accuracy enhancements in the new file-editing commands compared to the original ones.

\subsection{SWE-bench-lite}
\begin{table*}[!t]
    \centering
    \caption{Performance of \textsc{Infant Agent} on SWE-bench. All RAG and Agent results are from official SWE-bench leaderboard,  while the 0-shot results was self-implemented. We did not include agents with warnings from the leaderboard in the table.
  \includegraphics[height=0.7em]{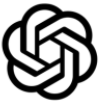}: OpenAI. \includegraphics[height=0.7em]{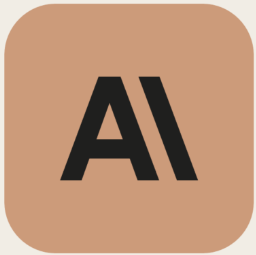}: Claude. \textcolor{red}{\faLock}: Close source. \cmark: Open source. Auto. Software Dev.: Automation software development. 3.5 S.: 3.5 Sonnet.\\}
    \label{tab:SWE_performance}
\begin{tabular}{cccccc}
\hline
\multicolumn{1}{l}{\textbf{Method}} & \textbf{Name}                            & \textbf{\% Resolved} & \textbf{Model Used}                                   & \textbf{Agent Scope} & \textbf{Open Source}     \\ \hline
\multirow{31}{*}{Agent}    & MarsCode Agent                           & $39.33$                & -                                                     & Code IDE             & \textcolor{red}{\faLock} \\
                           & Honeycomb                                & $38.33$                & -                                                     & -                    & \textcolor{red}{\faLock} \\
                           & Gru                                      & $35.67$                & -                                                     & Code IDE             & \textcolor{red}{\faLock} \\
                           & Isoform                                  & $35.00$                & -                                                     & Auto. Software Dev.  & \textcolor{red}{\faLock} \\
                           & SuperCoder2.0                            & $34.00$                & -                                                     & Auto. Software Dev.  & \textcolor{red}{\faLock} \\
                           & MarsCode Agent                           & $34.00$                & \includegraphics[width=0.3cm]{Icon/openai.png} 4o     & Code IDE             & \textcolor{red}{\faLock} \\
                           & Lingma Agent                             & $33.00$                & -                                                     & Fix Github issue     & \textcolor{red}{\faLock} \\
                           & AutoCodeRover                            & $30.67$                & \includegraphics[width=0.3cm]{Icon/openai.png} 4o     & Fix Github issue     & \cmark                   \\
                           & \cellcolor[HTML]{B7B7B7} \textsc{Infant Agent} & \cellcolor[HTML]{B7B7B7} $30.00$                & \cellcolor[HTML]{B7B7B7} \includegraphics[width=0.3cm]{Icon/openai.png} 4o     & \cellcolor[HTML]{B7B7B7} AI engineer          & \cellcolor[HTML]{B7B7B7} -                        \\
                           & Q Developer Agent                        & $29.67$                & -                                                     & AI engineer          & \textcolor{red}{\faLock} \\
                           & Agentless + RepoGraph                    & $29.67$                & \includegraphics[width=0.3cm]{Icon/openai.png} 4o     & Fix Github issue     & \cmark                   \\
                           & CodeR                                    & $28.33$                & \includegraphics[width=0.3cm]{Icon/openai.png} 4      & Fix Github issue     & \textcolor{red}{\faLock} \\
                           & MASAI                                    & $28.00$                & \includegraphics[width=0.3cm]{Icon/openai.png} 4o     & Fix Github issue     & \textcolor{red}{\faLock} \\
                           & SIMA                                     & $27.67$                & \includegraphics[width=0.3cm]{Icon/openai.png} 4o     & Fix Github issue     & \textcolor{red}{\faLock} \\
                           & Agentless                                & $27.33$                & \includegraphics[width=0.3cm]{Icon/openai.png} 4o     & Fix Github issue     & \cmark                   \\
                           & Moatless Tools                           & $26.67$                & \includegraphics[width=0.3cm]{Icon/claude.png} 3.5 S. & Fix Github issue     & \cmark                   \\
                           & OpenDevin                                & $26.67$                & \includegraphics[width=0.3cm]{Icon/claude.png} 3.5 S. & AI engineer          & \cmark                   \\
                           & Agent-101                                & $26.67$                & \includegraphics[width=0.3cm]{Icon/openai.png} 4o     & AI engineer          & \textcolor{red}{\faLock} \\
                           & Aider                                    & $26.33$                & -                                                     & AI engineer          & \textcolor{red}{\faLock} \\
                           & HyperAgent                               & $25.33$                & -                                                     & AI engineer          & \textcolor{red}{\faLock} \\
                           & Moatless Tools                           & $24.67$                & \includegraphics[width=0.3cm]{Icon/openai.png} 4o     & Fix Github issue     & \cmark                   \\
                           & IBM SWE-1.0                              & $23.67$                & -                                                     & AI engineer          & \textcolor{red}{\faLock} \\
                           & GenAgent                                 & $23.67$                & \includegraphics[width=0.3cm]{Icon/openai.png} 4      & AI engineer          & \textcolor{red}{\faLock} \\
                           & SWE-agent                                & $23.00$                & \includegraphics[width=0.3cm]{Icon/claude.png} 3.5 S. & AI engineer          & \cmark                   \\
                           & OpenDevin                                & $22.00$                & \includegraphics[width=0.3cm]{Icon/openai.png} 4o     & AI engineer          & \cmark                   \\
                           & Navie Agent                              & $21.67$                & \includegraphics[width=0.3cm]{Icon/openai.png} 4o     & Auto. Software Dev.  & \cmark                   \\
                           & AutoSE                                   & $21.67$                & \includegraphics[width=0.3cm]{Icon/openai.png} 4o     & -                    & \textcolor{red}{\faLock} \\
                           & Q Developer Agent                        & $20.33$                & \includegraphics[width=0.3cm]{Icon/openai.png} 4o     & AI engineer          & \textcolor{red}{\faLock} \\
                           & AutoCodeRover                            & $19.00$                & \includegraphics[width=0.3cm]{Icon/openai.png} 4      & Fix Github issue     & \textcolor{red}{\faLock} \\
                           & SWE-agent                                & $18.33$                & \includegraphics[width=0.3cm]{Icon/openai.png} 4o     & AI engineer          & \cmark                   \\
                           & SWE-agent                                & $18.00$                & \includegraphics[width=0.3cm]{Icon/openai.png} 4      & AI engineer          & \cmark                   \\ \hline
\multirow{3}{*}{RAG}       & -                                        & $4.33$                 & \includegraphics[width=0.3cm]{Icon/claude.png} 3 Opus                                         & -                    & -                        \\
                           & -                                        & $3.00$                 & \includegraphics[width=0.3cm]{Icon/claude.png} 2                                              & -                    & -                        \\
                           & -                                        & $2.67$                 & \includegraphics[width=0.3cm]{Icon/openai.png} 4                                                 & -                    & -                        \\ \hline
\multirow{2}{*}{0-shot}    & -                                        & $1.33$                 & \includegraphics[width=0.3cm]{Icon/claude.png} 3.5 S.                                     & -                    & -                        \\
                           & -                                        & $0.33$                 & \includegraphics[width=0.3cm]{Icon/openai.png} 4o                                               & -                    & -                        \\ \hline
                           &                                          &                      &                                                       &                      &                         
\end{tabular}
\end{table*}

\textbf{\textit{Dataset Description:}} 
SWE-bench~\cite{yang2024swe} is a dataset consisting of $2,294$ software engineering problems drawn from real GitHub issues and corresponding pull requests across 12 popular Python repositories. The input for this dataset is the description of a GitHub issue raised by a real user, and the agent is required to automatically generate a \texttt{Patch} file to resolve the GitHub issue. Since the API cost for testing the full SWE-bench dataset could be quite high, an official test subset, SWE-bench-lite, is provided.

\textbf{\textit{Experiment Setup:}} 
In the testing process, we initialized each level of agents with GPT-4o (temperature = $0.7$) and used evaluation conditions consistent with the official SWE-bench leaderboard: all submissions are Pass@1, do not use \texttt{hints\_text}, and are in the unassisted setting. The maximum number of iterations was set to 100, with up to 3 self-correction attempts. Automated linting prompts were enabled after code edits, the maximum timeout 
in the sandbox was set to $120$s, and the maximum cost per iteration was capped at 10 dollars.

\textbf{\textit{Experiment Analysis:}} 
The performance of \textsc{Infant Agent} on the SWE-bench is shown in Table~\ref{tab:SWE_performance}. 
Based on the data in Table 1, it is evident that most of the leading agent architectures remain closed-source, with limited technical details available. Among open-source agents, the performance of \textsc{Infant Agent} is only $0.67\%$ lower than AutoCodeRover, surpassing all other open-source agents.

When compared to specialized code-focused agents, such as MentatBot and AutoCodeRover, \textsc{Infant Agent} demonstrates a broader range of application scenarios, indicating its versatility beyond code-specific tasks. Additionally, when compared to generalized AI agents like OpenDevin and Aider, \textsc{Infant Agent} achieves superior accuracy. Using the 4o model for initialization, \textsc{Infant Agent} achieves an accuracy that is $8$ percentage points higher than OpenDevin.

The comparison between RAG models and agent-based systems shows a clear advantage for agents in SWE-bench performance. While RAG models can leverage retrieval for knowledge-intensive tasks, they lack the structured command execution and adaptability seen in agents. Agents, such as \textsc{Infant Agent}, perform significantly better due to their ability to manage complex workflows and apply task-specific actions, which RAG models are not equipped to handle effectively. This results in higher accuracy and more reliable task completion for agents, highlighting their superiority over RAG models for engineering and iterative problem-solving tasks.

This analysis highlights \textsc{Infant Agent}'s competitive performance, particularly among open-source and generalized AI agents, suggesting its effectiveness in both code-specific and broader engineering tasks.

\subsection{AIME \& Codeforce}
\label{exp:AIME_Codeforce}
\begin{table*}[t]
    \centering
    \caption{Performance of \textsc{Infant Agent} on AIME and Codeforce dataset. API-Cost are calculated based on OpenAI's pricing standards as of October 2024. The results for Claude 3.5 Sonnet on the AIME2024 dataset are sourced from their official documentation~\cite{Claude3_Model_Card}, while other results were obtained from our own testing. Maj: Majority voting. \textcolor{red}{\faLock}: Close Source. \\}
    \label{tab:AIME_Codeforce}
\begin{tabular}{ccccc}
\hline
Dataset                            & Model                   & prompting method  & Accuracy(\%) & API-Cost(\$) \\ \hline
\multirow{9}{*}{AIME2024}          & o-1 mini                & \textcolor{red}{\faLock}                 & $63$     &           $0.15$                \\
                                   & o-1 preview             & \textcolor{red}{\faLock}                 & $37$      &        $0.76$                  \\
                                   & \cellcolor[HTML]{B7B7B7} gpt-4o + Qwen-2.5 72B   & \cellcolor[HTML]{B7B7B7}Infant-AI Agent      & \cellcolor[HTML]{B7B7B7}$37$      & \cellcolor[HTML]{B7B7B7}$0.37$                     \\
                                   & Claude 3.5 Sonnet       & Maj@64 0-shot CoT & $27.60$   & -                        \\
                                   & gpt-4o                  & MACM~\cite{lei2024macm} Agent        & $26.70$   & $0.61$                      \\
                                   & gpt-4o                  & Maj@16 0-shot CoT & $16.70$   & $0.23$                      \\
                                   & Claude 3.5 Sonnet       & 0-shot CoT        & $16.00$   & -                        \\
                                   & gpt-4o                  & CoT               & $13.30$   & $0.06$                     \\
                                   & gpt-4o                  & 0-shot            & $13.30$   & $0.01$                     \\ \hline
\multirow{7}{*}{Codeforce} & o-1 mini                & \textcolor{red}{\faLock}                 & $36.60$   &            $0.07$            \\
                                   & o-1 preview             &  \textcolor{red}{\faLock}                 & $30.00$   &            $0.31$             \\
                                   &\cellcolor[HTML]{B7B7B7} gpt-4o + Qwen-2.5 72B   & \cellcolor[HTML]{B7B7B7}Infant-AI Agent      & \cellcolor[HTML]{B7B7B7}$26.70$   &      \cellcolor[HTML]{B7B7B7}   $0.03$                 \\
                                   & Claude 3.5 Sonnet & Maj@64 0-shot CoT &     $20.00$     &            -              \\
                                   & gpt-4o                  & CoT               & $16.70$   &            $0.03$              \\
                                   & Claude 3.5 Sonnet & 0-shot CoT        &    $16.70$      &             -             \\
                                   & gpt-4o                  & 0-shot            & $16.70$   &             $0.01$             \\ \hline
\end{tabular}
\end{table*}
\textbf{\textit{Dataset Description:}} 
The American Invitational Mathematics Examination (AIME) is the second exam in the series of exams used to challenge bright students on the path toward choosing the team that represents the United States at the International Mathematics Olympiad (IMO). 
To prevent data contamination, we used the o1 model testing standard and selected 2024 exam questions to test \textsc{Infant Agent}. 

Codeforces is a website that hosts competitive programming contests. 
Similarly, to prevent data contamination, we selected the most recent four Codeforces contests after the release of o1: contests 969 Div1, 970, 971, and 972 to test \textsc{Infant Agent}.

\textbf{\textit{Experiment Setup:}} 
During the testing phase, we initialized the brain-level agents with GPT-4o and the hand-level agents with Qwen2.5-72B-Instruct~\cite{hui2024qwen2}. 

Model setting: The temperature for GPT-4o was set to $0.7$, while inference for Qwen2.5 72B instruct was conducted using the \texttt{vllm} package with the following settings: $temperature=1.0$, $top\_p=1.0$, $top\_k=-1$, $tensor\_parallel\_size=8$, $kv\_cache\_allocation=0.95$, $max\_tokens=9632$, and $max\_retry=3$. 

Agent setting: $max\_iterations=100$, with up to $self\_correction\_times=3$, $sandbox\_timeout=120s$ , $max\_cost=\$10$. Results are uniformly recorded as Pass@1.

\textbf{\textit{Experiment Analysis:}} 
As shown in Table~\ref{tab:AIME_Codeforce}, we compared the accuracy of different models on these two datasets using various prompting methods and benchmarked them against the current state-of-the-art reasoning model, the o1 series. Results show that supported by the \textsc{Infant Agent} workflow, the combination of 4o and the open-source Qwen2.5-72B-Instruct achieves the same accuracy as o1-preview on the AIME2024 dataset, with nearly half the API cost. In addition, though o1-mini’s performance significantly surpasses other methods, \textsc{Infant Agent} still solved two problems that o1-mini could not: AIME-2024-I-15 and AIME-2024-II-14. This situation did not occur with other methods. On the Codeforces dataset, while its accuracy is slightly lower than o1-preview, the API cost is reduced by almost $90\%$.

\subsection{GPQA Diamond}
\label{exp:GPQA_Diamond}
\textbf{\textit{Dataset Description:}} 
The GPQA Diamond~\cite{rein2023gpqa} dataset contains $198$ PhD-level questions covering various fields, including Organic Chemistry, Quantum Mechanics, Astrophysics, Genetics, and more. Solving these problems requires the agent not only to have deep logical reasoning abilities but also to be able to retrieve information from the web and to write and execute code for scientific computations.

\textbf{\textit{Experiment Setup:}}
We initialized the brain-level agent of \textsc{Infant Agent} with GPT-4o and Claude-3.5-Sonnet, and consistently used Qwen2.5-72B-Instruct to initialize the hands-level agent. The \texttt{litellm} package was used to standardize the output format for both Claude-3.5-Sonnet and GPT-4o. All other experimental parameters were kept consistent with the setup used in Experiments~\ref{exp:AIME_Codeforce}. For the test data format, we used the \texttt{random} function to shuffle one correct answer and three incorrect answers, presenting them in choice format below the question description.

\textbf{\textit{Experiment Analysis:}} 
\begin{table}[H]
    \centering
    \caption{Performance of \textsc{Infant Agent} on the GPQA Diamond Dataset. 
\includegraphics[height=0.7em]{Icon/openai.png}: OpenAI. \includegraphics[height=0.7em]{Icon/claude.png}: Claude 3.5 Sonnet. \includegraphics[height=0.7em]{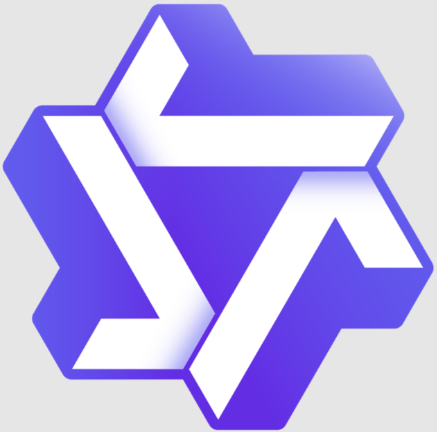}: Qwen2.5-72B-Instruct. \textcolor{red}{\faLock}: Close source. Maj: Majority voting.\\}
    \label{tab:GPQA_Diamond}
\begin{tabular}{ccc}
\hline
\textbf{Model}                      & \begin{tabular}[c]{@{}c@{}}\textbf{Prompting} \\ \textbf{Method}\end{tabular}& \begin{tabular}[c]{@{}c@{}}\textbf{Accuracy} \\ \textbf{ \%}\end{tabular} \\ \hline
\includegraphics[width=0.3cm]{Icon/openai.png} o1                         & \textcolor{red}{\faLock}                 & $78.0$       \\
\includegraphics[width=0.3cm]{Icon/openai.png} o1-preview                 & \textcolor{red}{\faLock}                 & $72.2$     \\
\cellcolor[HTML]{B7B7B7}\includegraphics[width=0.3cm]{Icon/claude.png} + \includegraphics[width=0.3cm]{Icon/Qwen2.5.png} & \cellcolor[HTML]{B7B7B7}\textsc{Infant Agent}      & \cellcolor[HTML]{B7B7B7}$71.7$     \\
Human Expert               & -                 & $69.7$     \\
\includegraphics[width=0.3cm]{Icon/claude.png}           & Maj@32+5-shot+CoT  & $67.2$     \\
\includegraphics[width=0.3cm]{Icon/claude.png}           & 0-shot+CoT        & $59.4$     \\
\cellcolor[HTML]{B7B7B7}\includegraphics[width=0.3cm]{Icon/openai.png} 4o + \includegraphics[width=0.3cm]{Icon/Qwen2.5.png}           & \cellcolor[HTML]{B7B7B7}\textsc{Infant Agent}      & \cellcolor[HTML]{B7B7B7}$58.0$       \\
\includegraphics[width=0.3cm]{Icon/openai.png} 4o                     & Maj@64            & $56.1$     \\
\includegraphics[width=0.3cm]{Icon/openai.png} 4o                     & 0-shot            & $50.0$       \\ \hline
\end{tabular}
\end{table}
In Table~\ref{tab:GPQA_Diamond}, results for o1 and human experts are drawn from the official o1 documentation~\cite{OpenAI_Learning_to_Reason}, while results for Claude 3.5 Sonnet + Maj@32+5-shot+CoT and Claude 3.5 Sonnet + 0-shot+CoT are sourced from Claude AI’s official report~\cite{anthropic2024claude}. Supported by the \textsc{Infant Agent} workflow, Claude 3.5 Sonnet + Qwen2.5 72B achieves an accuracy of $71.7\%$, surpassing human experts ($69.7\%$), without requiring additional fine-tuning of a critic model—a step OpenAI most likely includes in the o1 series~\cite{mcaleese2024llm,lightman2023let}, based on disclosed materials. Additionally, the Infant Agent workflow significantly outperforms other prompting approaches, demonstrating enhanced accuracy and resource efficiency, especially on challenging, high-complexity datasets.

\subsection{Error Command Correction Test}
\label{exp:Error_Command_Correction}
We tested the correction capability of the Hierarchical Agent Collaboration System for agent command misjudgments.

\textbf{\textit{Dataset Description:}} 
We selected a pure code task dataset, LiveCodeBench. It is a comprehensive and contamination-free evaluation benchmark for LLMs focused on code, which continuously gathers new problems over time. \textbf{In theory, this dataset requires only code generation, without any browser-related operations.}

\textbf{\textit{Experiment Setup:}} 
We selected two types of commands from the \textsc{Infant Agent} command library: file-editing commands and browser commands. To evaluate command generation accuracy, we compared the frequency of unintended browser command generation for this task using two distinct prompting methods, hierarchical prompting and flat prompting. In the flat prompting approach, we provided the model with a 1-shot example containing a mix of file-editing and browser commands. This analysis was conducted using both open-source and closed-source models to assess performance across different model types.
For closed-source models, we used GPT-4o, while for open-source models, we used Qwen2.5-72B-instruct. The LiveCodeBench timeframe was set from 1/1/2024 to 11/1/2024. During testing, all model temperatures were set to $0.0$, and results were recorded using pass@1 scores.

\textbf{\textit{Experiment Analysis:}} 
 In Table~\ref{tab:Command_misjudgments}, under the hierarchical prompting structure, the model did not generate any browser commands, whereas under the flat prompting structure, the model generated over $10\%$ browser commands. This misdirected the model's reasoning, leading to a decrease in accuracy. For Qwen2.5, the model's accuracy dropped by $10.7\%$, a level of precision loss that is unacceptable.
\begin{table}[H]
    \centering
    \caption{Command misjudgments correction capability of the Hierarchical prompting structure. Qwen2.5: Qwen2.5-72B-Instruct.\\}
    \label{tab:Command_misjudgments}
\begin{tabular}{cccc}
\hline
\begin{tabular}[c]{@{}c@{}}\textbf{Prompting}   \\ \textbf{structure}\end{tabular} & \textbf{Model}   & \begin{tabular}[c]{@{}c@{}}\textbf{Browser} \\ \textbf{ command}\end{tabular} & \begin{tabular}[c]{@{}c@{}}\textbf{Pass@1} \\ \textbf{\%}\end{tabular} \\ \hline
\cellcolor[HTML]{B7B7B7}Hierarchical                                                     & \cellcolor[HTML]{B7B7B7}GPT-4o  & \cellcolor[HTML]{B7B7B7}$0\%$\color{darkgreen}$\downarrow$                                                                & \cellcolor[HTML]{B7B7B7}$40.7$\color{darkgreen}$\uparrow$    \\
Flat                                                             & GPT-4o  & $11.70\%$                                                               & $34.8$   \\
\cellcolor[HTML]{B7B7B7}Hierarchical                                                     & \cellcolor[HTML]{B7B7B7}Qwen2.5 & \cellcolor[HTML]{B7B7B7}$0\%$\color{darkgreen}$\downarrow$                                                                & \cellcolor[HTML]{B7B7B7}$41.4$\color{darkgreen}$\uparrow$   \\
Flat                                                             & Qwen2.5 & $13.90\%$                                                               & $30.7$   \\ \hline
\end{tabular}
\end{table}

\subsection{API Token Savings from Memory Retrieval}
\label{exp:API_Token_Savings}
In this section, we compared the API token cost before and after memory retrieval and its impact on accuracy.
To ensure a sufficient level of task complexity, we selected $50$ test samples from SWE-bench-lite in which \textsc{Infant Agent} required over $70$ iterations. We then tested scenarios with and without memory retrieval. We keep all the experiment setups the same as the Experiment~\ref{exp:Error_Command_Correction}.
\begin{table}[H]
    \centering
    \caption{Impact of Memory Retrieval on Task Resolution and API Cost. \includegraphics[width=0.3cm]{Icon/openai.png}:GPT-4o. \includegraphics[width=0.3cm]{Icon/claude.png}:Claude-3.5-Sonnet. \includegraphics[width=0.3cm]{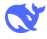}: DeepSeek-Coder-V2-236B. \includegraphics[width=0.2cm]{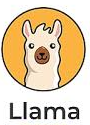}:Llama-3.1-70B-Instruct. \textcolor{red}{\ding{55}}:Without memory retrieval. \textcolor{darkgreen}{\ding{51}}:With memory retrieval. \\}
    \label{tab:Memory_retrieve}
\begin{tabular}{ccccc}
\hline
\begin{tabular}[c]{@{}c@{}}\textbf{Memory}   \\ \textbf{Retrieval}\end{tabular} & \begin{tabular}[c]{@{}c@{}}\textbf{Brain}   \\ \textbf{Agent}\end{tabular} & \begin{tabular}[c]{@{}c@{}}\textbf{Hands}   \\ \textbf{Agent}\end{tabular} & \textbf{\% Resolved} & \begin{tabular}[c]{@{}c@{}}\textbf{API}   \\ \textbf{Cost(\$)}\end{tabular} \\ \hline
\textcolor{red}{\ding{55}}                & \includegraphics[width=0.3cm]{Icon/openai.png}                & \includegraphics[width=0.3cm]{Icon/openai.png}                & $12$          & $7.81$                   \\
\textcolor{darkgreen}{\ding{51}}              & \includegraphics[width=0.3cm]{Icon/openai.png}                & \includegraphics[width=0.3cm]{Icon/Deepseek.png}          & $13$          & $2.13$                   \\
\textcolor{darkgreen}{\ding{51}}               & \includegraphics[width=0.3cm]{Icon/openai.png}                & \includegraphics[width=0.3cm]{Icon/Qwen2.5.png}              & $12$          & $2.03$                   \\
\textcolor{darkgreen}{\ding{51}}               & \includegraphics[width=0.3cm]{Icon/openai.png}                & \includegraphics[height=0.7em]{Icon/LLama.png}          & $8$           & $2.17$                   \\ \hline
\textcolor{red}{\ding{55}}              & \includegraphics[width=0.3cm]{Icon/claude.png} & \includegraphics[width=0.3cm]{Icon/claude.png} & $14$          & $4.42$                   \\
\textcolor{darkgreen}{\ding{51}}                & \includegraphics[width=0.3cm]{Icon/claude.png} & \includegraphics[width=0.3cm]{Icon/Deepseek.png}          & $14$          & $0.92$                   \\
\textcolor{darkgreen}{\ding{51}}                & \includegraphics[width=0.3cm]{Icon/claude.png} & \includegraphics[width=0.3cm]{Icon/Qwen2.5.png}              & $14$          & $0.99$                   \\
\textcolor{darkgreen}{\ding{51}}                & \includegraphics[width=0.3cm]{Icon/claude.png} & \includegraphics[height=0.7em]{Icon/LLama.png}          & $12$          & $0.82$                   \\ \hline
\end{tabular}
\end{table}

Table~\ref{tab:Memory_retrieve} shows that the memory retrieval mechanism has a significant impact on API costs. Without memory retrieval, the API cost is relatively high. For example, GPT-4o has a cost of \$$7.81$ without memory retrieval, which drops substantially to as low as \$$2.03$ with memory retrieval enabled. Similarly, for the Claude model, the cost is \$$4.42$ without memory retrieval and decreases to a minimum of \$$0.82$ when it is enabled.

In terms of task resolution rate (\% Resolved), memory retrieval has a limited impact on the solution accuracy, with resolution rates remaining relatively stable across different configurations. However, enabling memory retrieval can significantly reduce costs while maintaining similar accuracy across various models and agent configurations. This indicates that the memory retrieval mechanism can effectively optimize resource usage and reduce API costs without significantly affecting task resolution performance.
\subsection{File Editing Accuracy Improvement}
\label{exp:File_Editing}
When executing file-editing commands, the agent must accurately generate line numbers; otherwise, misalignment errors may occur, leading to incorrect edits. In this section, we tested the error-correction capability of our new file-editing command compared to the original SWE-Agent file-editing command.

\begin{figure}[ht]
    \centering
\includegraphics[width=0.48\textwidth]{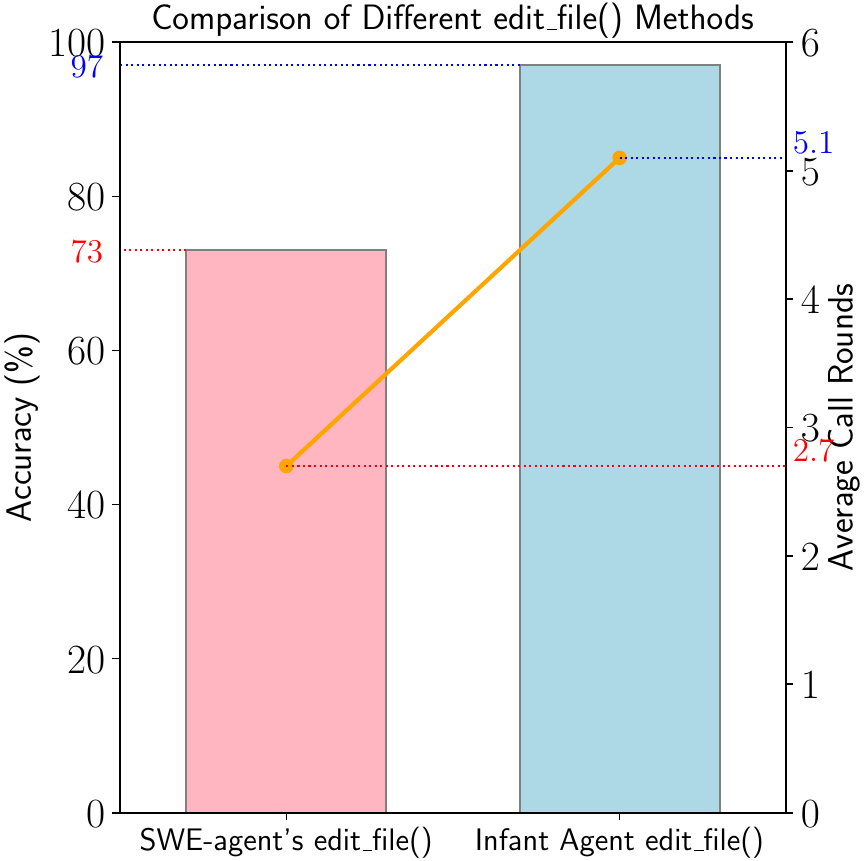} 
    \caption{Comparison of accuracy and average call rounds between SWE-Agent edit\_file() and \textsc{Infant Agent} edit\_file() command.} 
    \label{fig:edit_file_result} 
\end{figure}

We selected the same 50 test cases as in Experiment~\ref{exp:API_Token_Savings}. However, this time we focused specifically on all file-editing commands within these cases. Since sequential file edits do not inherently trigger errors, we manually reviewed a total of 351 file-editing commands across these 50 test cases. The experimental results are shown in Figure~\ref{fig:edit_file_result}.

\textbf{\textit{Experiment Analysis:}} The results in the Figure~\ref{fig:edit_file_result} demonstrate that the Infant Agent \texttt{edit\_file()} method achieves a substantial improvement in accuracy at the cost of a slight increase in average call rounds. Specifically, the Infant Agent method reached an impressive $97\%$ accuracy compared to $73\%$ for the original SWE-Agent method, highlighting a significant enhancement in generating precise line numbers and content for file edits, which reduces the occurrence of sequencing errors and incorrect edits. Although the Infant Agent method required an average of $5.1$ call rounds compared to $2.7$ for SWE-Agent, this trade-off in additional calls enables a marked increase in accuracy, making Infant Agent more effective and reliable for file editing tasks.
\section{Conclusion}
We present three major contributions: \textsc{Infant Agent}, an advanced agent that can perform deep logical reasoning, invoke tools, and engage in self-reflection; a hierarchical agent collaboration system to address output inefficiencies caused by an excessive number of built-in commands or overly lengthy few-shot examples; and a memory retrieval mechanism, which reduces API token costs by $80\%$ compared to using the full memory for each inference, thereby optimizing resource efficiency. Together, these innovations significantly enhance the Infant Agent's adaptability, cost-effectiveness, and capability to handle complex tasks.

\section{Future Work}
\begin{enumerate}[label=\textbf{\arabic*.}]
    \item We plan to expand the current Agent framework from a text modality to a multimodal one. 
    Two potential technical approaches: 1. Moving the mouse at the pixel level~\cite{BrowserGym}, 2. First performing image parsing, then locating~\cite{kang2025segvg,kang2024robin3d}.
    \item Train a File-Editing model.
    \item Verify step by step and enhance GPT's error-correction capability through reinforcement learning~\cite{lightman2023let}.
    \item Teach model how to use tools instead of using long prompts~\cite{lei2024autocoder}.
\end{enumerate}




\nocite{langley00}

\bibliography{citation}
\bibliographystyle{mlsys2025}

\appendix

\section{Pipeline Demonstration Example}
\label{Appendix:Pipeline_Demonstration_Example}
To illustrate the \textbf{actual} operational logic of Infant Agent, we selected the following example and included terminal screenshots from the program’s runtime to demonstrate Infant Agent's workflow:

First, the user submits a query to the agent:
\begin{figure}[H]
    \centering
\includegraphics[width=0.48\textwidth]{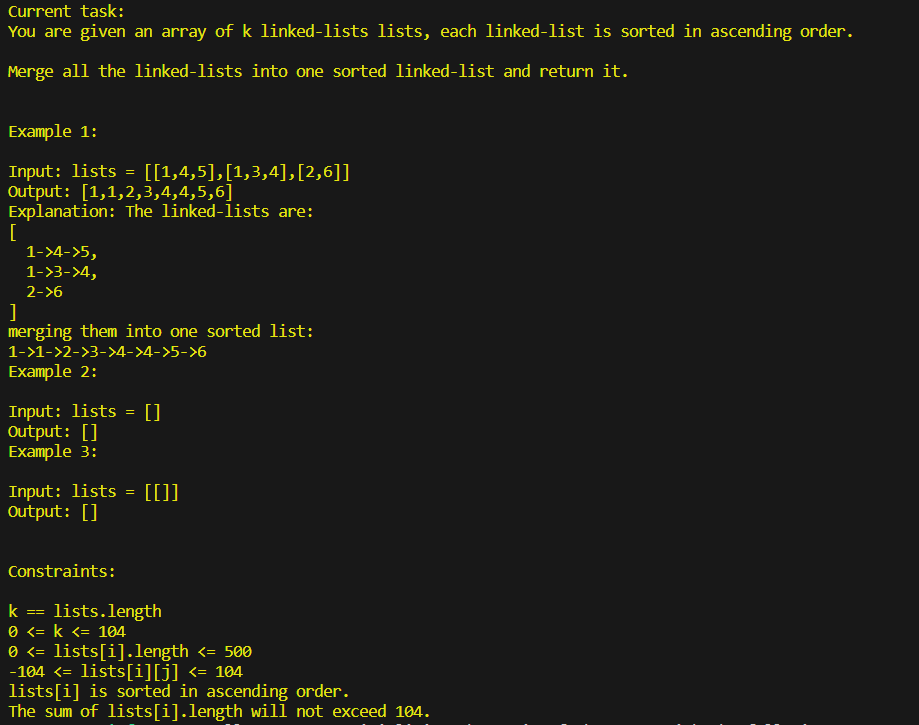} 
    \caption{User Input.}  
\end{figure}

The first turn analysis begin:
\begin{figure}[H]
    \centering
\includegraphics[width=0.48\textwidth]{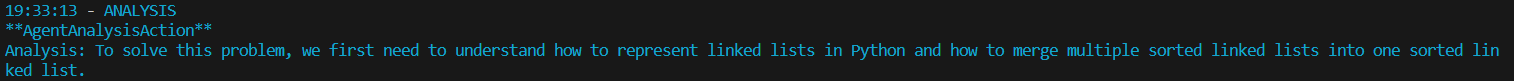} 
    \caption{Analysis one.}  
\end{figure}

\begin{figure}[H]
    \centering
\includegraphics[width=0.48\textwidth]{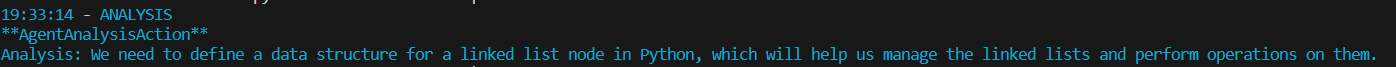} 
    \caption{Analysis two.}  
\end{figure}

The first turn Task begin:
\begin{figure}[H]
    \centering
\includegraphics[width=0.48\textwidth]{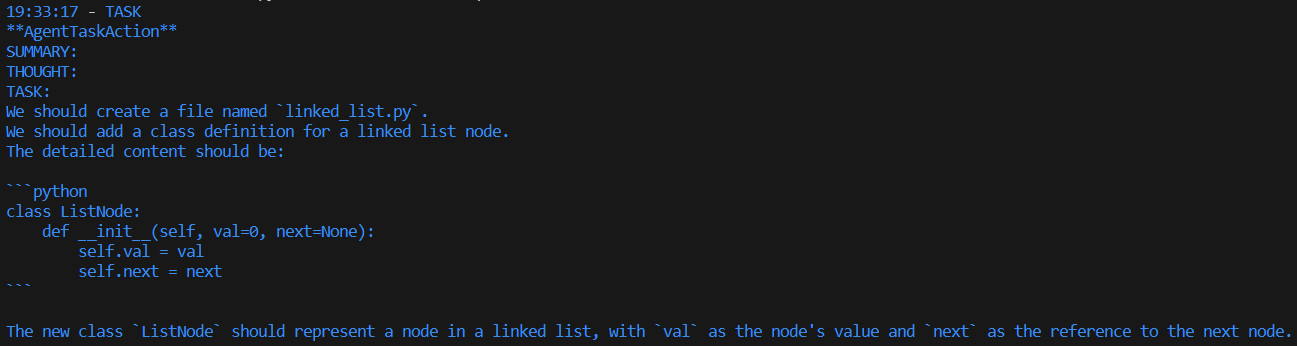} 
    \caption{Task one.}  
\end{figure}

The first turn execution begin:
\begin{figure}[H]
    \centering
\includegraphics[width=0.48\textwidth]{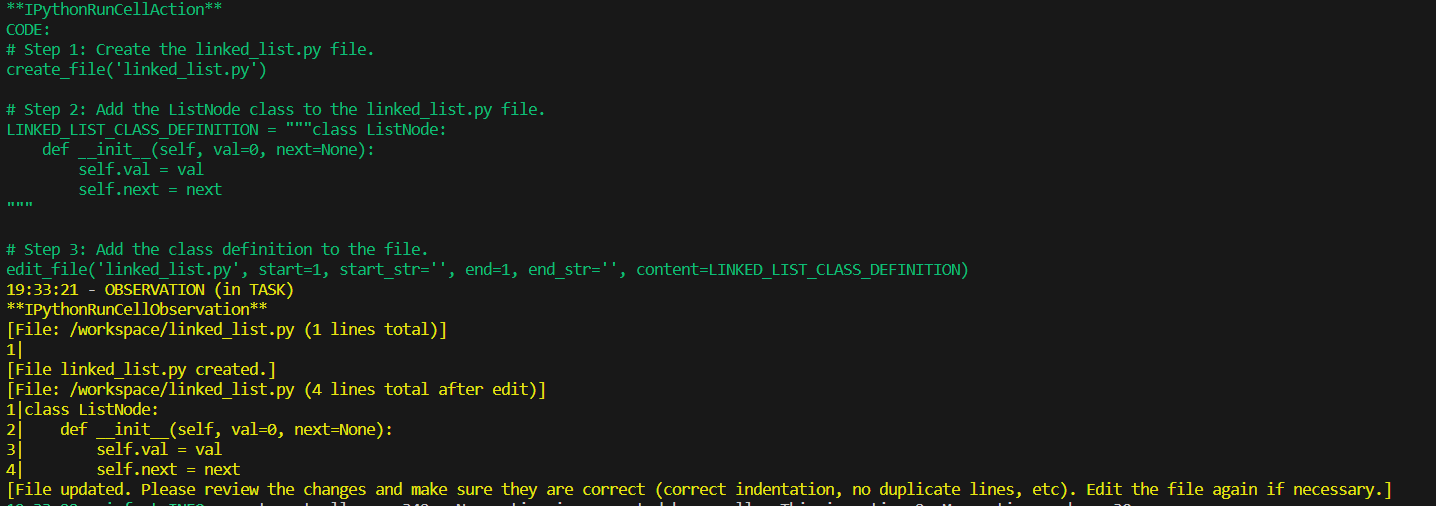} 
    \caption{Execution one.}  
\end{figure}

It passed the evaluation and begin summary:
\begin{figure}[H]
    \centering
\includegraphics[width=0.48\textwidth]{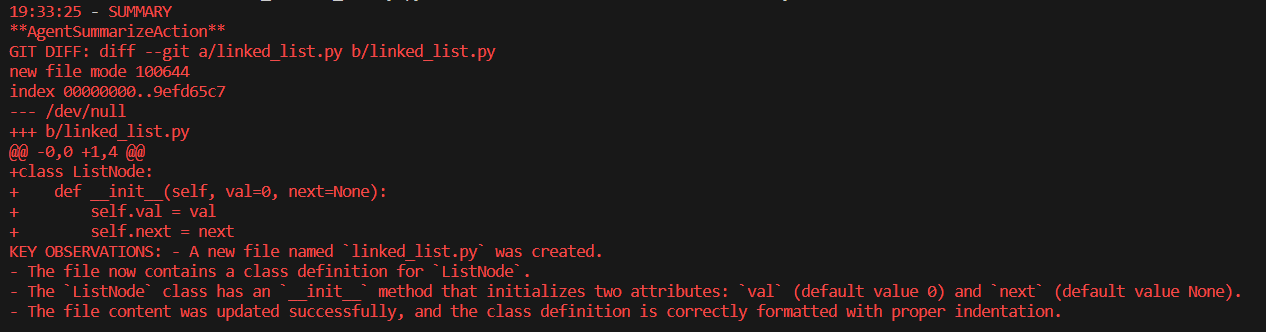} 
    \caption{Summary one.}  
\end{figure}

The second turn analysis begin:
\begin{figure}[H]
    \centering
\includegraphics[width=0.48\textwidth]{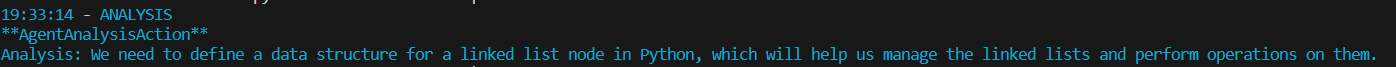} 
    \caption{Analysis two.}  
\end{figure}

\begin{figure}[H]
    \centering
\includegraphics[width=0.48\textwidth]{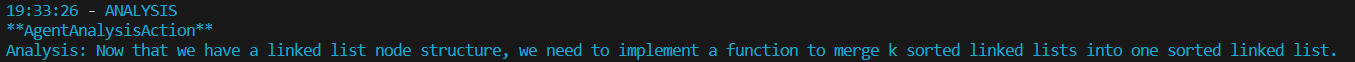} 
    \caption{Analysis three.}  
\end{figure}

The second turn task begin:
\begin{figure}[H]
    \centering
\includegraphics[width=0.48\textwidth]{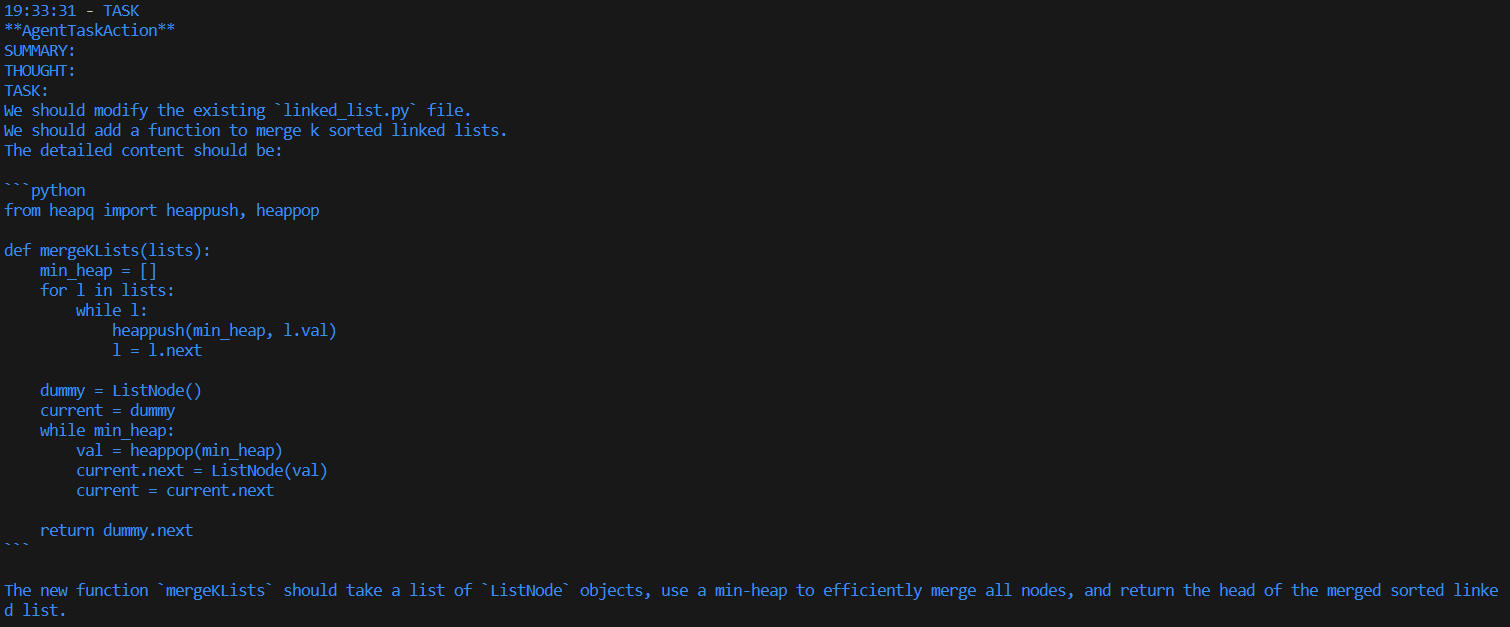} 
    \caption{Task two.}  
\end{figure}

The second turn execution begin:
\begin{figure}[H]
    \centering
\includegraphics[width=0.48\textwidth]{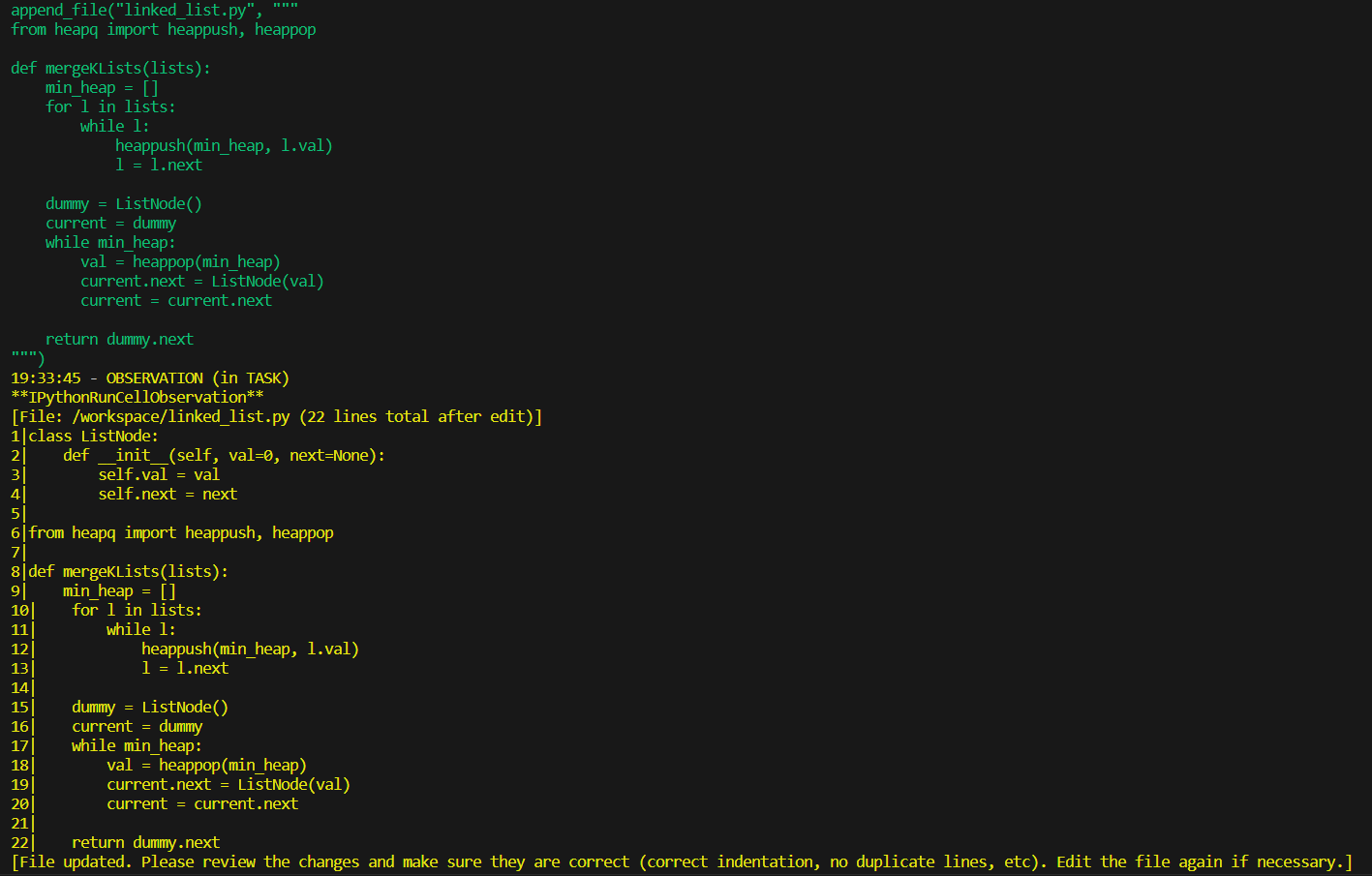} 
    \caption{Execution two.}  
\end{figure}

The second turn summary begin:
\begin{figure}[H]
    \centering
\includegraphics[width=0.48\textwidth]{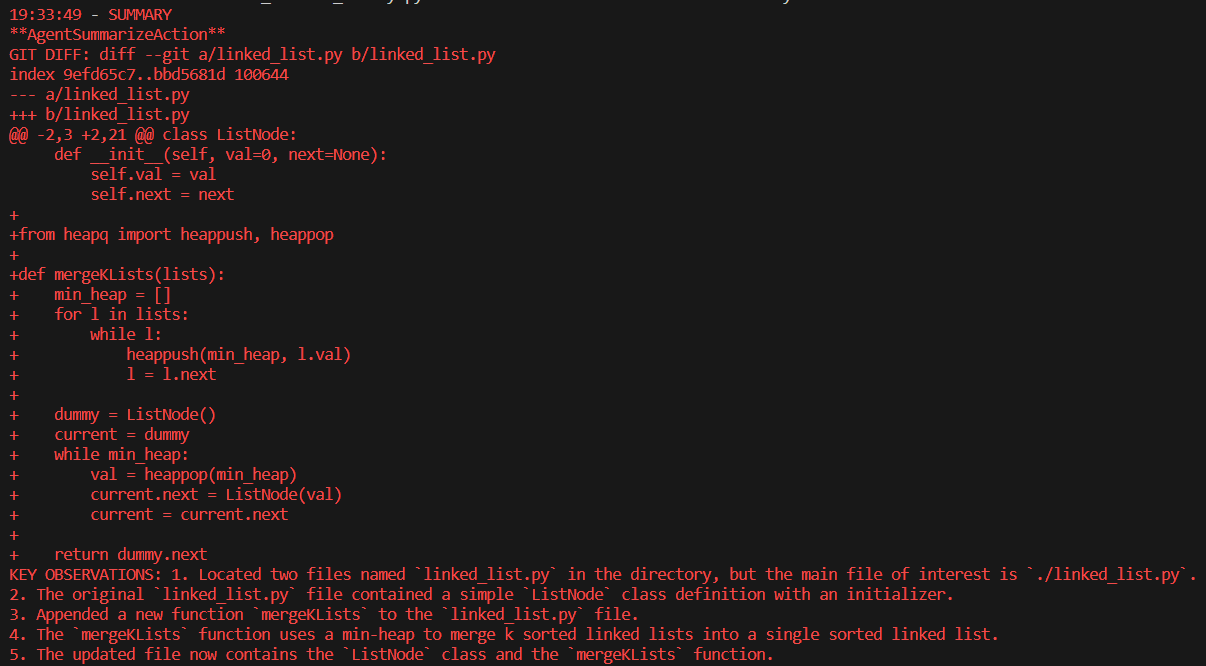} 
    \caption{Summary two.}  
\end{figure}

The third turn task begin:
\begin{figure}[H]
    \centering
\includegraphics[width=0.48\textwidth]{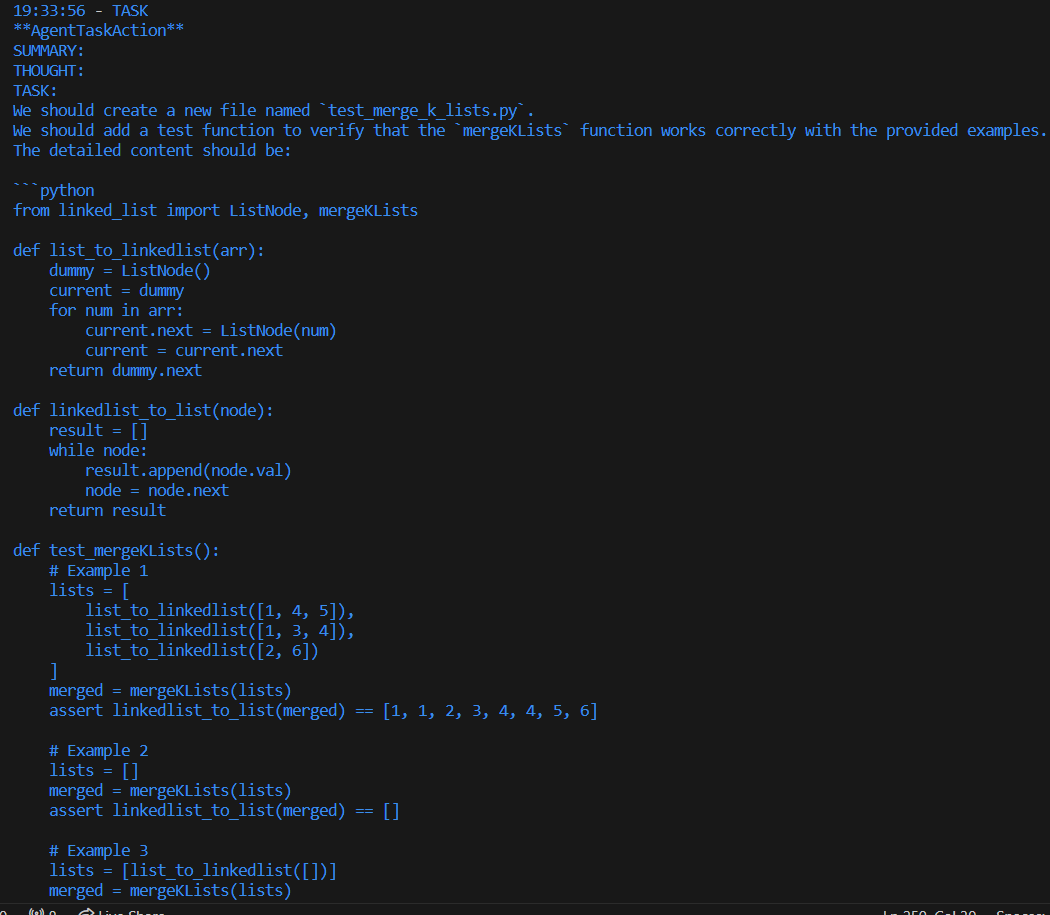} 
    \caption{Task three.}  
\end{figure}

The third turn task begin:
\begin{figure}[H]
    \centering
\includegraphics[width=0.48\textwidth]{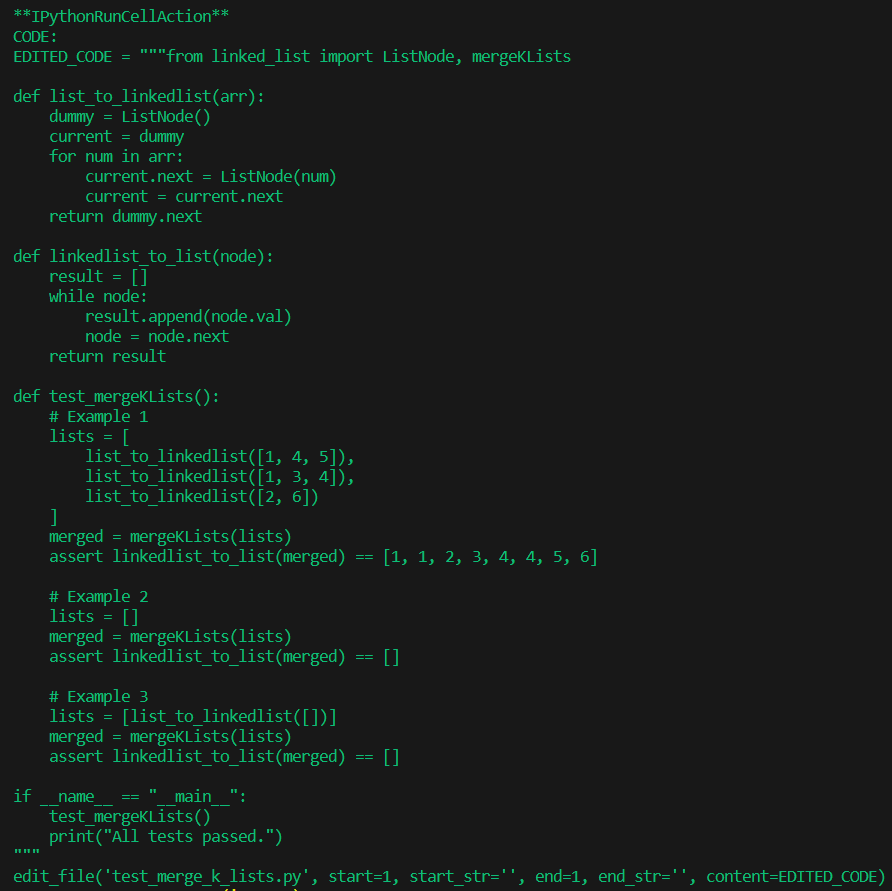} 
    \caption{Execution three.}  
\end{figure}

\begin{figure}[H]
    \centering
\includegraphics[width=0.48\textwidth]{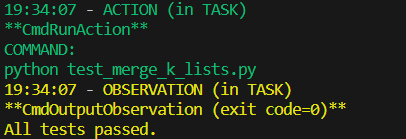} 
    \caption{Execution four.}  
\end{figure}

Task finished:
\begin{figure}[H]
    \centering
\includegraphics[width=0.48\textwidth]{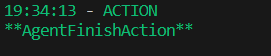} 
    \caption{Execution four.}  
\end{figure}

%

\end{document}